\documentclass[11pt]{article}

\usepackage[final]{acl}

\usepackage{times}
\usepackage{latexsym}
\usepackage{multirow}

\usepackage[T1]{fontenc}

\usepackage[utf8]{inputenc}

\usepackage{microtype}

\usepackage{inconsolata}

\usepackage{graphicx}

\usepackage{pgfplots}
\usepackage[most]{tcolorbox}
\usepackage{enumitem}
\usepackage{verbatim}

\usepackage{adjustbox}
\usepackage[utf8]{inputenc}
\usepackage{amsmath}
\usepackage{algorithm}
\usepackage{algorithmicx} 
\usepackage{algpseudocode}
\usepackage{natbib}
\usepackage{float} 
\usepackage{graphicx}  
\usepackage{booktabs}  
\usepackage{siunitx}   
\usepackage[section]{placeins}
\usepackage{tabularx,booktabs,pifont}
\usepackage{booktabs}
\usepackage{siunitx}
\usepackage[most]{tcolorbox}

\newcommand{\cmark}{\ding{51}} 

\title{MemORAI: Memory Organization and Retrieval via Adaptive Graph Intelligence for LLM Conversational Agents}

\author{
  \textbf{Hung Pham Van\textsuperscript{1}\footnotemark[1]},
  \textbf{Nguyen Manh Hieu\textsuperscript{1}\footnotemark[1]},
  \textbf{Khang Pham Tran Tuan\textsuperscript{2}\footnotemark[1]}, \\
  \textbf{Nam Le Hai\textsuperscript{2,\dag}},
  \textbf{Linh Ngo Van\textsuperscript{2}}, 
  \textbf{Diep Thi-Ngoc Nguyen\textsuperscript{3}},
  \textbf{Trung Le\textsuperscript{4}}
  \bigskip \\
  \textsuperscript{1}Independent Researcher, \textsuperscript{2}Hanoi University of Science and Technology, \\
  \textsuperscript{3}VNU University of Engineering and Technology,  \textsuperscript{4}Monash University
}
\begin{document}

\maketitle
\renewcommand{\thefootnote}{\fnsymbol{footnote}}
\footnotetext[1]{Equal contribution}
\footnotetext[2]{Corresponding author: \href{mailto:namlh@soict.hust.edu.vn}{namlh@soict.hust.edu.vn}}
\renewcommand{\thefootnote}{\arabic{footnote}}

\begin{abstract}
Large Language Models (LLMs) lack persistent memory for long-term personalized conversations. Existing graph-based memory systems suffer from information dilution, absent provenance tracking, and uniform retrieval that ignores query context. We introduce MemORAI (Memory Organization and Retrieval via Adaptive Graph Intelligence), a framework that integrates three innovations: selective memory filtering with dual-layer compression to retain user-persona-relevant content, a provenance-enriched multi-relational graph tracking factual origins at the turn level, and query-adaptive subgraph retrieval with Dynamic Weighted PageRank that applies query-conditioned edge weighting. Evaluated on \texttt{LOCOMO} and \texttt{LongMemEval} benchmarks, MemORAI achieves state-of-the-art performance in memory retrieval and personalized response generation, demonstrating that selective storage, enriched representation, and adaptive retrieval are essential for coherent, personalized LLM agents.
\end{abstract}

\section{Introduction}
Human cognition relies on a dynamic memory system that balances acquisition, consolidation, and retrieval to sustain personalized interactions without cognitive overload~\citep{liu2025advanceschallengesfoundationagents}. Large Language Models (LLMs), despite excelling in reasoning and generation~\citep{geminiteam2025geminifamilyhighlycapable, grattafiori2024llama3herdmodels, yang2025qwen3technicalreport, deepseekai2025deepseekv3technicalreport}, lack this persistence. Constrained by limited context windows, they lose crucial details~\citep{liu2023lostmiddlelanguagemodels} and reset to a stateless baseline across sessions~\citep{timoneda2025memoryneedtestingmodel, yuan2024personalizedlargelanguagemodel}, making ephemeral prompting a fragile substitute that amplifies hallucinations~\citep{lewis2021retrievalaugmentedgenerationknowledgeintensivenlp}.

Memory-augmented approaches address this through external stores and selective retrieval~\citep{liu2025advanceschallengesfoundationagents, Wang_2024}. While retrieval-augmented generation (RAG)~\citep{lewis2021retrievalaugmentedgenerationknowledgeintensivenlp} and vector-based systems~\citep{yuan2024personalizedlargelanguagemodel, pan2025memoryconstructionretrievalpersonalized, tan2025prospectretrospectreflectivememory} have advanced factual grounding, they struggle with relational and temporal structures~\citep{Wang_2024}. Graph-based representations offer richer modeling through interconnected entities and relations~\citep{chhikara2025mem0buildingproductionreadyai, gutirrez2025ragmemorynonparametriccontinual}, yet existing systems reveal critical gaps: hierarchical methods like RAPTOR require expensive re-clustering~\citep{sarthi2024raptorrecursiveabstractiveprocessing}, sparse graphs like Mem0g bias toward high-degree nodes~\citep{chhikara2025mem0buildingproductionreadyai}, and sophisticated approaches like HippoRAG 2 propagate scores uniformly without query-conditioned adaptation~\citep{gutirrez2025ragmemorynonparametriccontinual}. Crucially, no existing system filters user-persona-relevant content from generic dialogue or tracks provenance at the turn level, leading to information dilution and opacity in factual origins.

To address these limitations, we introduce \textbf{MemORAI—Memory Organization and Retrieval via Adaptive Graph Intelligence}—a framework that integrates selective memory filtering, provenance-enriched graph construction, and query-adaptive retrieval with dynamic edge weighting. Our contributions are:
\begin{itemize}[leftmargin=*]
\item \textbf{Selective Memory Filtering:} A memory gate that retains only user-persona-relevant content while generating segment-level summaries to preserve global context, improving storage efficiency and retrieval precision.
\item \textbf{Provenance-Enriched Knowledge Graph:} A heterogeneous graph architecture with entity, turn, and segment nodes featuring explicit turn-level provenance tracking for transparent auditing and fine-grained retrieval.
\item \textbf{Dynamic Weighted PageRank:} A query-adaptive retrieval method that constructs focused subgraphs through multi-aspect search and applies query-conditioned edge weighting to prioritize semantically aligned evidence.
\end{itemize}

\section{Related Work}

\paragraph{Memory Granularity.}
Early retrieval-based memory systems segmented dialogue history at either the turn or session level~\citep{yuan2024personalizedlargelanguagemodel, Wang_2024}. While turn-level units preserve fine details, they fragment context; session-level aggregation, by contrast, introduces irrelevant noise. To balance coherence and relevance, \citet{pan2025memoryconstructionretrievalpersonalized} proposed \textbf{SECOM}, which segments dialogue into coherent topical units and applies compression-based denoising, while \citet{xu2025amemagenticmemoryllm} developed \textbf{A-MEM}, constructing dynamic ``atomic notes'' linked by shared attributes. These works highlight that fixed granularity constrains both retrieval efficiency and adaptability, motivating structures capable of hierarchical and relational reasoning beyond flat memory units.

\paragraph{Structured Memory Representations.}
Beyond flat chunking, hierarchical and graph-based memories enable relational reasoning and associative recall. \textbf{RAPTOR}~\citep{sarthi2024raptorrecursiveabstractiveprocessing} recursively clusters and summarizes text into a multi-level tree, supporting thematic retrieval but at significant computational cost due to recursive LLM summarization and full re-clustering during updates. \textbf{Mem0g}~\citep{chhikara2025mem0buildingproductionreadyai} introduces a graph-based memory representing conversational knowledge as entity–relation triplets, facilitating multi-hop reasoning but limited by shallow semantics—nodes often store only surface names without entity descriptions, and synonym edges are defined by name similarity rather than conceptual meaning. Similarly, \textbf{HippoRAG 2}~\citep{gutirrez2025ragmemorynonparametriccontinual} employs Personalized PageRank over dense-sparse knowledge graphs for continual retrieval, yet its propagation remains uniform across edges and its synonym linking depends solely on lexical overlap between entity names. These simplifications cause brittle relational inference, synonym noise, and uniform ranking insensitive to query semantics.

\paragraph{Adaptive Retrieval.}
Recent studies explore dynamic retrieval mechanisms to enhance contextual sensitivity. \textbf{Reflective Memory Management (RMM)}~\citep{tan2025prospectretrospectreflectivememory} refines memory organization via prospective and retrospective reflection, using reinforcement feedback to adapt retrieval weights. \textbf{HippoRAG 2} extends this direction through query-conditioned Personalized PageRank, but without relation-level semantic modulation. Consequently, existing methods remain constrained by fixed propagation rules and limited personalization, often conflating high-degree node connectivity with relevance. While recent efforts in Graph RAG have begun to mitigate this structural bias by incorporating multi-aspect semantic reranking \citep{hieu2025magix}, they typically apply this after traversal. MemORAI, in contrast, directly embeds query-conditioned semantic modulation into the traversal process itself.

In contrast, MemORAI addresses these limitations through three key mechanisms. First, selective memory filtering with dual-layer compression tackles information dilution by retaining only user-persona-relevant content while preserving global context through segment summaries. Second, provenance-enriched graph construction enables transparent auditing by tracking factual origins at the turn level—a capability absent in prior work. Third, query-adaptive subgraph retrieval with Dynamic Weighted PageRank overcomes uniform propagation by applying query-conditioned edge weighting, enabling context-sensitive retrieval without exhaustive graph traversal. Together, these mechanisms establish a cohesive memory lifecycle that integrates selective storage, enriched representation, and adaptive retrieval.

\section{Methodology}
\begin{figure*}[t]
    \centering
    \includegraphics[width=\textwidth]{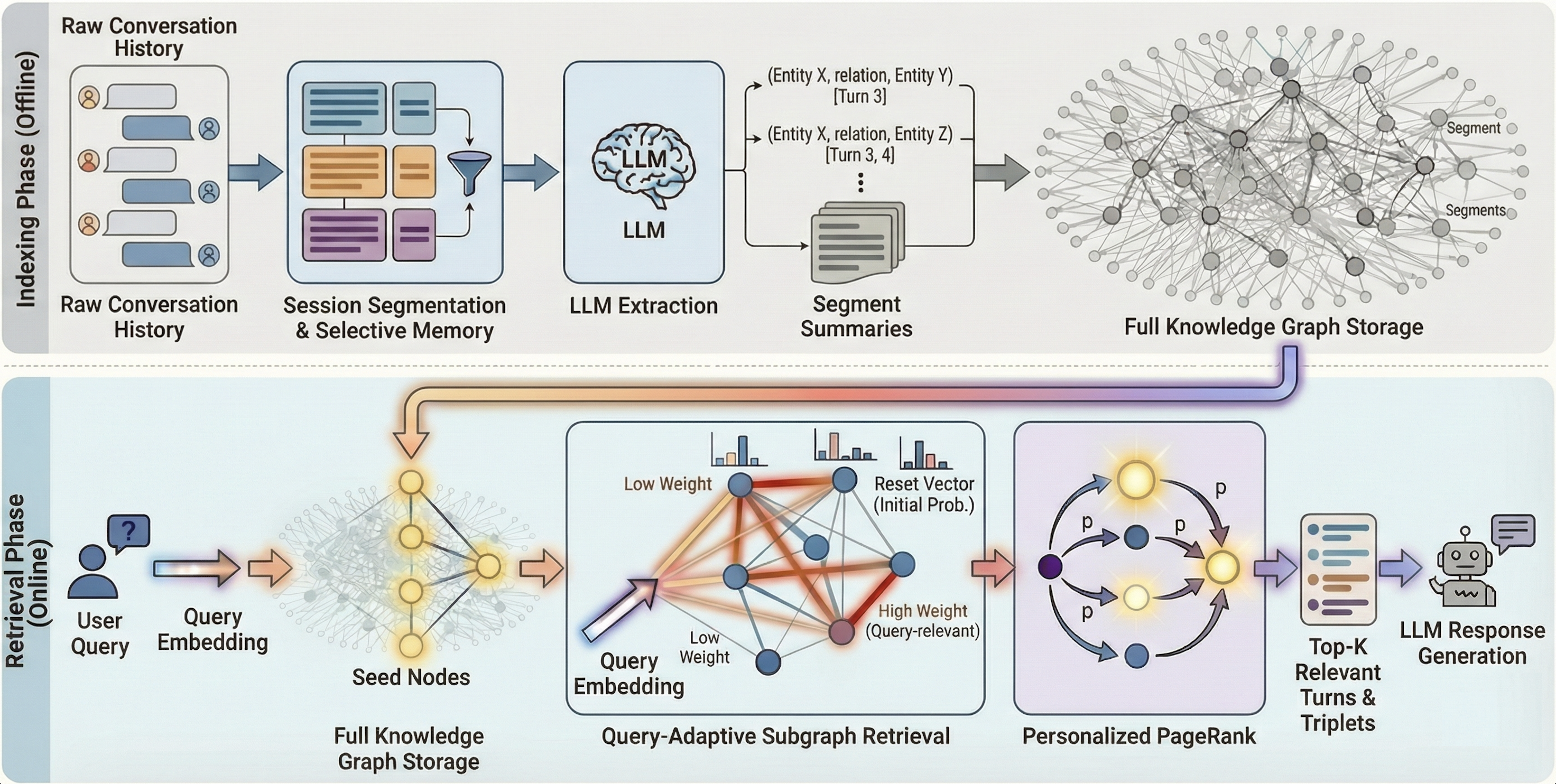}
\caption{Overview of MemORAI's three-phase pipeline. (1) \textbf{Session Segmentation and Selective Compression}: Conversations are segmented topically; a memory gate retains user-relevant utterances and summarizes discarded content. (2) \textbf{Provenance-Enriched Graph Construction}: LLM extraction produces entity-relation triplets with turn-level provenance in a heterogeneous graph. (3) \textbf{Query-Adaptive Retrieval and Generation}: Multi-aspect search identifies seed nodes; query-focused subgraph assembly enables Dynamic Weighted PageRank with query-conditioned edge weights; top-ranked turns and triplets guide personalized response generation.}
\label{fig:overview}
\end{figure*}

\textbf{MemORAI} implements a streamlined three-phase pipeline for long-term personalized dialogue agents (Figure~\ref{fig:overview}): (1) \textbf{Session Segmentation and Selective Compression}—dialogues are segmented topically, and a memory gate retains only user-relevant utterances while summarizing generic discourse for coherence (\S\ref{sec:segmentation}); (2) \textbf{Provenance-Enriched Graph Construction}—entity-relation triplets are extracted from retained messages and embedded in a heterogeneous graph of entities, turns, and segments with explicit turn-level provenance (\S\ref{sec:graph}); (3) \textbf{Query-Adaptive Retrieval and Generation}—multi-aspect retrieval seeds a query-focused subgraph, Dynamic Weighted PageRank ranks nodes by query-conditioned semantic alignment, and top-ranked turns with supporting triplets are formatted into provenance-aware prompts for personalized response generation (\S\ref{sec:retrieval}).

\subsection{Session Segmentation \& Selective Compression}
\label{sec:segmentation}
Following SECOM~\citep{pan2025memoryconstructionretrievalpersonalized}, we first decompose raw multi-session conversations into semantically coherent segments $S_i = \{t_1, t_2, \dots, t_m\}$ via LLM prompting (Appendix~\ref{app:segmentation}). For each segment, we apply a selective memory gate that identifies and retains only messages containing user-specific episodic content—personal facts, preferences, commitments, and identity markers—producing a filtered set $M_i \subseteq S_i$ (Appendix~\ref{app:selective_filtering}). 

To preserve global context from discarded messages, we generate a segment-level summary $\sigma_i$ via LLM prompting (Appendix~\ref{app:segment_summary}). This dual-layer approach stores both $M_i$ (fine-grained personal content) and $\sigma_i$ (global contextual anchor), preventing information loss while dramatically reducing storage overhead and filtering out noise. By focusing on memory-relevant content, this selective compression ensures that subsequent graph extraction and construction operate on high-quality, user-centric signals rather than generic conversational clutter.

\subsection{Provenance-Enriched Graph Construction}
\label{sec:graph}
From each filtered segment $M_i$, we construct a multi-relational knowledge graph $G = (V, E)$ with explicit provenance tracking. All components entities, descriptions, and triplets are extracted via LLM prompts that enforce turn-level citation (Figure~\ref{fig:overview}, prompts in Appendices~\ref{app:entity_extraction}~\&~\ref{app:triplet_extraction}).

\paragraph{Node Types.}
The graph includes three node types:
\begin{itemize}[leftmargin=*,topsep=3pt,itemsep=2pt]
\item \textit{Entity nodes} $e \in V_E$ store a \texttt{name}, fine-grained natural-language \texttt{description} (e.g., ``Alex---software engineer at XYZ, prefers async communication'')—an approach shown to preserve semantic details better than uniform node summaries \citep{hieu2025magix}—and \texttt{turn\_ids} for provenance. 
\item \textit{Turn nodes} $\tau \in V_T$ store \texttt{text}, \texttt{segment\_id}, and \texttt{turn\_id}. 
\item \textit{Segment nodes} $s \in V_S$ store summary $\sigma_i$ and \texttt{segment\_id}. 

Embeddings are computed from descriptions, turn text, and summaries respectively.
\end{itemize}

\paragraph{Edge Types.}
The graph includes three edge types:
\begin{itemize}[leftmargin=*,topsep=3pt,itemsep=2pt]
\item \textit{Entity-relation-entity edges} $e_1 \xrightarrow{r} e_2$ connect entities via typed relations (e.g., $(\text{Alex}, \text{works\_at}, \text{XYZ})$), storing \texttt{source\_turns} for turn-level provenance. 
\item \textit{Entity-turn edges} $e \leftrightarrow \tau$ link entities to their mentions. 
\item \textit{Turn-segment edges} $\tau \leftrightarrow s$ preserve dialogue hierarchy. 
\end{itemize}

This heterogeneous structure enables multi-hop reasoning and precise provenance-aware retrieval.

\subsection{Query-Adaptive Subgraph Retrieval \& Ranking}
\label{sec:retrieval}
Given a user query $q$, MemORAI retrieves relevant memory through a two-step process: query-focused subgraph retrieval via multi-aspect seeding, followed by dynamic weighted ranking.

\subsubsection{Query-Focused Subgraph Retrieval}

Unlike HippoRAG 2~\citep{gutirrez2025ragmemorynonparametriccontinual}, which applies ranking across the entire memory graph, we dynamically retrieve a sparse, query-focused subgraph $G_q = (V_q, E_q)$ at query time. Through multi-aspect parallel retrieval, we first identify top-$k$ seed nodes—both segment nodes (via summary embeddings) and entity nodes (via description embeddings) and top-$k$ relation edges (via triplet description embeddings) using semantic similarity search. We then perform one-hop neighborhood expansion from these seeds to include all directly connected turns, entities, and segments. This query-adaptive subgraph retrieval filters out irrelevant memory regions before ranking, reducing noise that could otherwise degrade ranking performance while preserving high-quality, contextually relevant evidence with full provenance links.

\subsubsection{Dynamic Weighted PageRank}

\begin{figure*}[t]
    \centering
    \includegraphics[width=\textwidth]{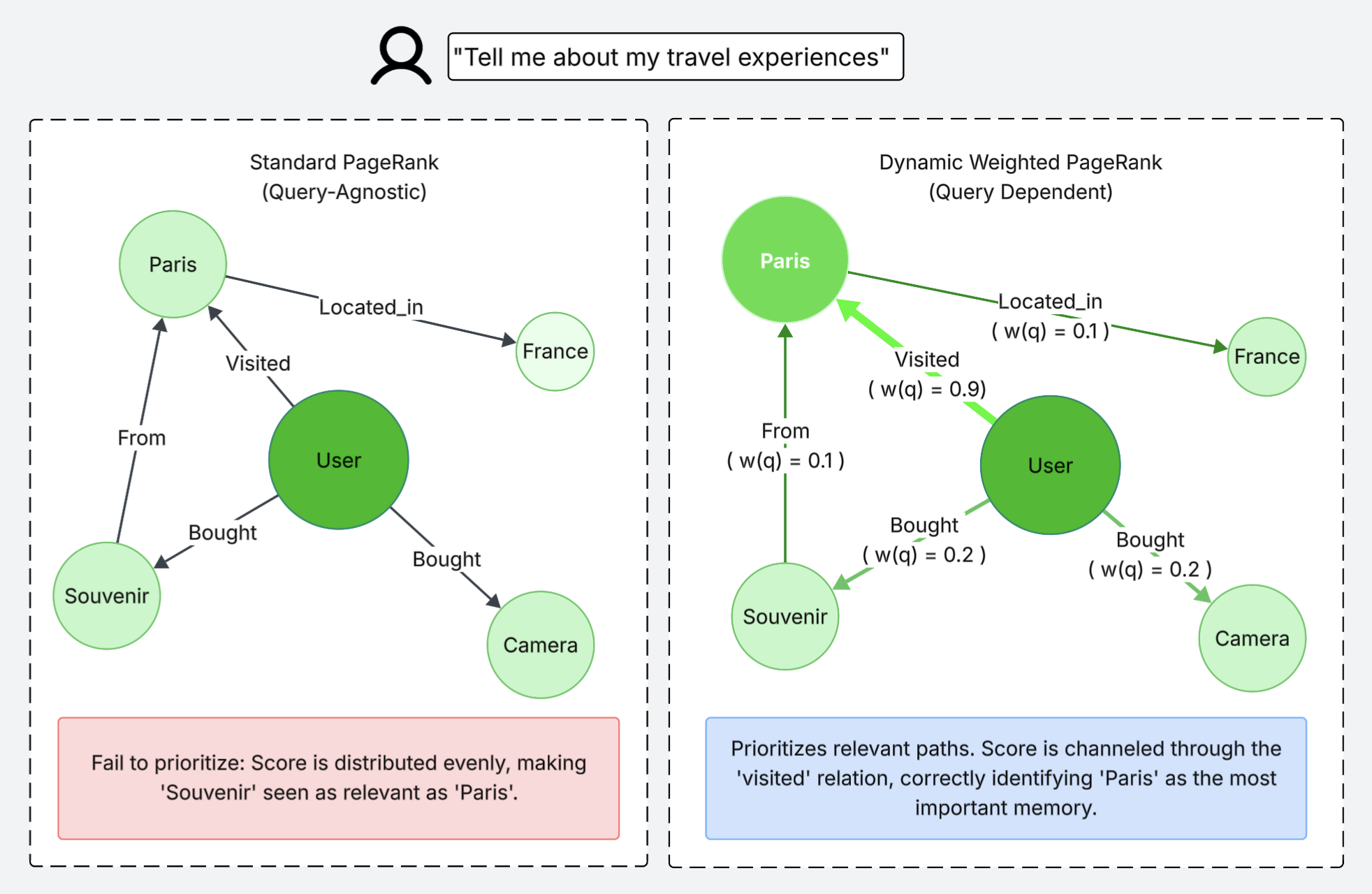}
    \caption{Traditional PageRank vs Dynamic Weighted PageRank}
    \label{fig:pagerank}
\end{figure*}

Traditional PageRank algorithms prioritize nodes with many high-quality neighbors, as scores propagate recursively from authoritative sources. HippoRAG 2~\citep{gutirrez2025ragmemorynonparametriccontinual} applies Personalized PageRank (PPR) with seed nodes extracted from queries and reset probabilities biased toward relevant starting points, enabling multi-hop reasoning through random walks over the knowledge graph. However, this uniform propagation mechanism can bias rankings toward nodes with dominant neighbor counts, potentially suppressing memory content in less-connected nodes that are nonetheless semantically relevant to the query. To address this limitation, our \textbf{Dynamic Weighted PageRank (DW-PR)} modulates score propagation based on query-conditioned edge weights that reflect semantic alignment rather than structural connectivity alone (see Figure~\ref{fig:pagerank}). The necessity of shifting from uniform to importance-aware weighting mirrors successful strategies in recent LLM alignment and cross-tokenizer distillation, where prioritizing highly informative signals over uniform processing yields superior performance across various optimization tasks \citep{truong2026ctpd, vu2026dwa, le2025token}. For each edge type, we define:
\begin{equation}
\small
w(u \to v) = \begin{cases}
    \text{sim}(q, e.\text{desc}), & u=e, v=\tau \\
    \text{sim}(q, r.\text{desc}), & u \xrightarrow{r} v \\
    \frac{1}{|\tau|} \sum_{e \in \tau} \text{sim}(q, e.\text{desc}), & u=\tau, v=s
\end{cases}
\label{eq:edge_weights}
\end{equation}
where $\text{sim}(\cdot, \cdot)$ denotes cosine similarity between query and description embeddings. All nodes in the subgraph are initialized with $\text{seed}(v)$ equal to their semantic similarity to $q$. DW-PR scores then propagate iteratively via:
\begin{equation}
\label{eq:pagerank}
    \text{PR}_{t+1}(v) = (1 - d) \cdot \text{seed}(v) + d \cdot S(v),
\end{equation}
where
\begin{equation}
\label{eq:pagerank-S}
    S(v) = \sum_{u \to v} \frac{w(u \to v)}{\sum_{u \to *} w(u \to *)} \text{PR}_t(u),
\end{equation}
and $d$ is the damping factor. This query-adaptive weighting ensures that semantically relevant but sparsely connected nodes can rank highly, preventing structural bias from overshadowing contextually critical memory content. 

After convergence, turn nodes $\tau$ are ranked by their final PageRank scores, and the top-$m$ turns are retrieved. For each retrieved turn, we also include all entity-relation triplets that cite it (i.e., where $\tau \in \text{source\_turns}$). These turns and supporting triplets are then formatted into a provenance-aware prompt that augments the conversational context, enabling the LLM to generate responses grounded in personalized memory with explicit citation of supporting evidence.

\section{Experiments}

\subsection{Experimental Settings}
\paragraph{Datasets \& Metrics.}
We evaluate on two long-horizon conversational memory benchmarks: \textbf{LongMemEval-s}~\citep{ICLR2025_d813d324}, and \textbf{LoCoMo-10} \cite{maharana2024evaluatinglongtermconversationalmemory}. These datasets target agent memory over sustained, multi-session dialogues rather than single-turn recall. Following an inference-only setting (no additional training or fine-tuning), we treat every QA pair in each benchmark as test data.
For \textbf{retrieval evaluation}, we report Recall@$k$ ($k \in \{3, 5, 10\}$) at both session-level and turn-level granularity. For \textbf{generation evaluation}, we measure both lexical and semantic fidelity using \textbf{F1}, \textbf{BLEU}, \textbf{ROUGE} (R1, R2, RL), and \textbf{BERTScore} \cite{zhang2020bertscoreevaluatingtextgeneration}. We additionally employ \textbf{GPT-4o} as a judge (GPT4o-J) to assess answer correctness on a normalized scale.

\paragraph{Baselines.}
We compare our method against a diverse set of approaches spanning full-history context, dense retrieval, memory-centric conversation, and structured RAG.
(1) \textbf{Full History}: uses the complete conversation records without explicit retrieval, accommodating up to a 128k-token window.
\textbf{Dense retrieval models}: (2) \emph{MPNet} \citep{song2020mpnetmaskedpermutedpretraining}, (3) \emph{Contriever} \citep{izacard2022unsuperviseddenseinformationretrieval}, (4) \emph{BGE-M3}, and (5) \emph{BM25}.
\textbf{Memory-based conversational models}: (6) \emph{LLM-RSum} \citep{wang2025recursivelysummarizingenableslongterm}, which recursively summarizes and updates a compact memory buffer;
(7) \emph{MPC} \citep{Lee_2023}, which leverages a pre-trained LLM to curate high-quality conversational memories;
(8) \emph{SeCom} \citep{pan2025memoryconstructionretrievalpersonalized}, which segments long dialogues into coherent topics with compression-based denoising;
(9) \emph{MemGAS}, which combines memory gating with adaptive summarization.
\textbf{Structured RAG models}: (10) \emph{HippoRAG 2} \citep{gutirrez2025ragmemorynonparametriccontinual}, which integrates knowledge-graph indexing with graph traversal;
(11) \emph{RAPTOR} \citep{sarthi2024raptorrecursiveabstractiveprocessing}, which applies recursive summarization and hierarchical clustering;
(12) \emph{LightRAG} \citep{guo2024lightrag} a lightweight graph-based retrieval approach; and
(13) \emph{MemTree} \citep{rezazadeh2024isolated}, which organizes memories in a hierarchical tree structure.

\paragraph{Implementation Details.}
We employ \texttt{openai/gpt-oss-20b} (decoding temperature $0$) uniformly across all modules and adopt top-$k=3$ retrieval. Memory embeddings are generated with \texttt{Contriever} to ensure fair comparison with prior work—not owing to its embedding quality, but to isolate the contribution of our design and algorithmic innovations.

\begin{table*}[t]
\centering
\small
\caption{QA performance on LongMemEval-s and LOCOMO-10. GPT4o-J denotes GPT-4o judge scores (\%). Best results in \textbf{bold}, second-best \underline{underlined}. RAPTOR returns hierarchical summaries rather than verbatim excerpts.}
\label{tab:qa-main}
\begin{tabular}{l|ccccccc}
\toprule
\multicolumn{8}{c}{\textbf{LongMemEval-s}} \\
\midrule
\textbf{Model} & GPT4o-J & F1 & BLEU & R-1 & R-2 & R-L & BERTScore \\
\midrule
Full History & 50.60 & 11.48 & 1.40 & 12.10 & 5.47 & 10.85 & 83.07 \\
BM25 & 42.00 & 14.19 & 4.30 & 22.08 & 11.10 & 21.37 & 86.53 \\
BGE-M3 & 47.60 & 12.30 & 3.82 & 19.19 & 8.80 & 18.56 & 86.14 \\
MPNet & 41.20 & 16.06 & 5.90 & 24.92 & 12.53 & 24.20 & 87.12 \\
Contriever & 41.00 & 23.94 & 9.25 & 32.10 & 15.50 & 30.63 & 88.40 \\
LLM-RSum & 35.40 & 12.29 & 2.09 & 13.01 & 5.55 & 11.52 & 83.60 \\
MPC & 53.80 & 13.60 & 1.74 & 14.27 & 6.49 & 12.95 & 83.49 \\
SeCom & 45.69 & \underline{29.13} & \underline{11.53} & \underline{36.91} & \underline{20.10} & \underline{35.83} & \underline{89.22} \\
HippoRAG 2 & 57.60 & 14.73 & 2.15 & 15.30 & 7.36 & 13.83 & 83.86 \\
RAPTOR & 32.20 & 12.08 & 1.90 & 12.73 & 5.82 & 11.25 & 83.50 \\
MemGAS & \underline{60.20} & 20.38 & 4.22 & 21.05 & 10.47 & 19.47 & 85.21 \\
LightRAG & 56.21 & 2.17 & 0.43 & 3.06 & 1.18 & 2.57 & 79.63 \\
MemTree & 21.80 & 8.57 & 2.27 & 12.02 & 4.38 & 10.44 & 83.49 \\
\midrule
\textbf{MemOrai (Ours)} & \textbf{75.55} & \textbf{45.99} & \textbf{11.54} & \textbf{50.63} & \textbf{25.25} & \textbf{50.02} & \textbf{90.37} \\
\bottomrule
\end{tabular}

\vspace{0.3cm}

\begin{tabular}{l|ccccccc}
\toprule
\multicolumn{8}{c}{\textbf{LOCOMO-10}} \\
\midrule
\textbf{Model} & GPT4o-J & F1 & BLEU & R-1 & R-2 & R-L & BERTScore \\
\midrule
Full History & 33.43 & 12.23 & 1.84 & 12.70 & 5.66 & 11.73 & 84.07 \\
BM25 & 28.05 & 14.19 & 4.30 & 22.08 & 11.10 & 21.36 & 86.53 \\
BGE-M3 & 28.80 & 12.30 & 3.82 & 19.19 & 8.80 & 18.56 & 86.14 \\
MPNet & 32.93 & 16.05 & 5.90 & 24.92 & 15.29 & 24.19 & 87.12 \\
Contriever & 32.15 & 14.64 & 5.03 & 22.94 & 11.12 & 22.29 & 86.77 \\
LLM-RSum & 22.56 & 9.14 & 0.99 & 9.82 & 3.38 & 8.98 & 83.45 \\
MPC & 40.38 & 14.81 & 1.99 & 15.10 & 6.83 & 14.13 & 84.43 \\
SeCom & 43.81 & \underline{21.33} & \underline{8.83} & \underline{34.09} & \underline{18.54} & \underline{33.02} & \underline{88.42} \\
HippoRAG 2 & 45.62 & 16.66 & 2.91 & 17.01 & 8.27 & 15.93 & 84.88 \\
RAPTOR & 31.72 & 14.55 & 2.88 & 15.09 & 7.49 & 14.18 & 84.48 \\
MemGAS & 41.07 & 17.66 & 3.61 & 18.00 & 8.93 & 16.99 & 85.13 \\
LightRAG & \underline{48.80} & 1.28 & 0.18 & 1.72 & 0.65 & 1.58 & 79.23 \\
Amem & 35.25 & 15.19 & 5.61 & 23.84 & 11.84 & 23.01 & 87.02 \\
Mem0 & 20.09 & 2.23 & 0.21 & 2.48 & 2.51 & 2.34 & 83.80 \\
MemTree & 29.44 & 9.57 & 1.67 & 14.01 & 5.06 & 12.91 & 84.41 \\
\midrule
\textbf{MemOrai (Ours)} & \textbf{60.22} & \textbf{56.71} & \textbf{33.00} & \textbf{57.90} & \textbf{42.57} & \textbf{56.58} & \textbf{91.71} \\
\bottomrule
\end{tabular}
\end{table*}

\subsection{Main Results}
\label{sec:main_results}

We report end-to-end QA performance (Table~\ref{tab:qa-main}), along with session- and turn-level retrieval results (Tables~\ref{tab:session-retrieval} and~\ref{tab:turn-retrieval}). All models—including ours—use \texttt{Contriever} for memory embeddings and \texttt{gpt-oss-20b} for generation, ensuring a controlled comparison.

\paragraph{Long context and standard retrieval show limited gains:}
The full-history baseline achieves moderate judge scores (50.60\% on LongMemEval-s), but performance drops notably on the more fragmented LOCOMO-10 (33.43\%). Dense retrievers (e.g., Contriever, BGE-M3) improve lexical metrics—Contriever reaches R-1 = 32.10 on LongMemEval-s—but their GPT-4o scores remain below 42\%, suggesting that embedding-based similarity alone may not reliably surface semantically relevant evidence for complex, multi-session questions.

\paragraph{Compression and static graphs face granularity trade-offs:}
Methods that compress dialogue (e.g., LLM-RSum, RAPTOR) show reduced lexical performance, possibly due to loss of fine-grained details. Graph-based approaches such as HippoRAG 2 perform well at session-level retrieval (75.53 R@3; Table~\ref{tab:session-retrieval}) but exhibit substantially lower turn-level recall (27.80 R@3; Table~\ref{tab:turn-retrieval}), indicating that coarse structural representations may not preserve sufficient turn-level provenance for precise QA.

\paragraph{Embedding quality does not fully explain performance:}
BGE-M3, which uses its own stronger embeddings, attains the highest turn-level recall among dense retrievers (67.97 R@3; Table~\ref{tab:turn-retrieval}) but achieves only 47.60\% GPT-4o-J (Table~\ref{tab:qa-main})—lower than several memory-centric methods (e.g., MemGAS: 60.20\%). This suggests that high embedding quality, while helpful, may not be sufficient without mechanisms for selective retention and contextualized retrieval.

\paragraph{Our approach shows consistent improvements across metrics:}
Using the same \texttt{Contriever} embeddings as baselines, MemORAI achieves the highest recall at both granularities (90.17 R@3 session, 71.13 R@3 turn on LongMemEval-s; Tables~\ref{tab:session-retrieval},~\ref{tab:turn-retrieval}) and the highest GPT-4o scores (75.55\% and 60.22\%). Notably, it outperforms BGE-M3 in turn-level retrieval (71.13 vs.\ 67.97) and judge score (75.55\% vs.\ 47.60\%) despite the latter's stronger embeddings. These results suggest that the proposed components—selective memory filtering, provenance-aware graph construction, and query-adaptive ranking—may help bridge the gap between retrieval precision and generation fidelity. Further ablation studies (§\ref{sec:ablation}) examine their individual contributions.
\begin{table*}[t]
\centering
\caption{Session-level retrieval performance. All methods use the same retrieval architecture. Best results in \textbf{bold}, second-best \underline{underlined}.}
\label{tab:session-retrieval}
\begin{tabular}{l|ccc|ccc}
\toprule
\textbf{Model} & \multicolumn{3}{c|}{\textbf{LongMemEval-s}} & \multicolumn{3}{c}{\textbf{LOCOMO-10}} \\
 & \textbf{Recall@3} & \textbf{Recall@5} & \textbf{Recall@10} & \textbf{Recall@3} & \textbf{Recall@5} & \textbf{Recall@10} \\
\midrule
MPNet & 66.17 & 76.38 & 85.11 & 45.92 & 53.98 & 68.58 \\
Contriever & 71.06 & 81.28 & 90.00 & 49.90 & 58.26 & 71.80 \\
LLM-RSum & 67.23 & 79.79 & 87.66 & 47.23 & 59.01 & 74.97 \\
MPC & 60.00 & 68.09 & 80.00 & 49.50 & 57.45 & 71.85 \\
SeCom & 71.06 & 80.43 & 89.15 & 53.86 & 62.01 & 73.44 \\
HippoRAG 2 & \underline{75.53} & 84.68 & 91.28 & 56.60 & 65.06 & 78.05 \\
MemGAS & 78.51 & \underline{88.94} & \underline{94.47} & \underline{57.30} & \underline{67.32} & \underline{81.82} \\
\textbf{MemOrai (Ours)} & \textbf{90.17} & \textbf{94.56} & \textbf{98.54} & \textbf{72.05} & \textbf{81.63} & \textbf{92.03} \\
\bottomrule
\end{tabular}
\end{table*}

\begin{table*}[t]
\centering
\caption{Turn-level retrieval performance. All methods use the same retrieval architecture. Best results in \textbf{bold}, second-best \underline{underlined}.}
\label{tab:turn-retrieval}
\begin{tabular}{l|ccc|ccc}
\toprule
\textbf{Model} & \multicolumn{3}{c|}{\textbf{LongMemEval-s}} & \multicolumn{3}{c}{\textbf{LOCOMO-10}} \\
 & \textbf{Recall@3} & \textbf{Recall@5} & \textbf{Recall@10} & \textbf{Recall@3} & \textbf{Recall@5} & \textbf{Recall@10} \\
\midrule
MPNet & 48.25 & 57.58 & 70.53 & 25.36 & 32.40 & 43.15 \\
Contriever & 45.07 & 57.02 & 69.40 & 24.36 & 30.30 & 37.75 \\
BGE-M3 & \underline{67.97} & \underline{79.36} & \underline{88.14} & 18.03 & 22.63 & 29.52 \\
BM25 & 56.72 & 64.68 & 73.69 & 26.05 & 32.10 & 39.21 \\
HippoRAG 2 & 27.80 & 38.85 & 53.90 & \underline{42.43} & \textbf{52.49} & \underline{61.81} \\
LightRAG & 13.41 & 23.86 & 48.66 & 8.33 & 13.90 & 26.03 \\
\textbf{MemOrai (Ours)} & \textbf{71.13} & \textbf{82.64} & \textbf{91.63} & \textbf{42.63} & \underline{51.97} & \textbf{64.68} \\
\bottomrule
\end{tabular}
\end{table*}

\subsection{Ablation Study}
\label{sec:ablation}

We conduct controlled ablations to assess the impact of each core component. Unless otherwise noted, all variants use the same \texttt{Contriever} embeddings and \texttt{gpt-oss-20b} backbone.

\subsubsection{Selective Memory Filtering and Topic Segmentation}
\label{sec:ablation_filtering}

Table~\ref{tab:ablation_segment} evaluates the role of topic segmentation and selective memory filtering. Removing topic segmentation leads to substantial performance degradation---e.g., turn-level R@10 drops from 91.63 to 23.86 on LongMemEval-s, and from 64.68 to 27.61 on LOCOMO-10. In contrast, ablating selective filtering has a more moderate impact (e.g., $-17.78$ on LongMemEval-s turn R@10), suggesting that segmentation provides a stronger structural prior for memory organization.

\begin{table}[H]
\centering
\caption{Topic Segmentation \& Selective Memory}
\label{tab:ablation_segment}
\small
\setlength{\tabcolsep}{4pt}
\renewcommand{\arraystretch}{1.05}
\begin{tabular}{l|cc}
\hline
\textbf{Method} & \textbf{R@10 Turn} & \textbf{R@10 Session} \\
\hline
\multicolumn{3}{c}{\textit{Locomo}} \\
\hline
Current (full) & 64.68 & 92.03  \\
w/o Selective  & 57.32 & 87.44  \\
w/o Topic Seg  & 27.61 & 66.02  \\
\hline
\multicolumn{3}{c}{\textit{Longmem\_s}} \\
\hline
Current (full) & 91.63 & 98.54  \\
w/o Selective  & 73.85 & 93.72  \\
w/o Topic Seg  & 23.86 & 75.79  \\
\hline
\end{tabular}
\end{table}

\begin{figure}[H]
    \centering
    \includegraphics[width=\columnwidth]{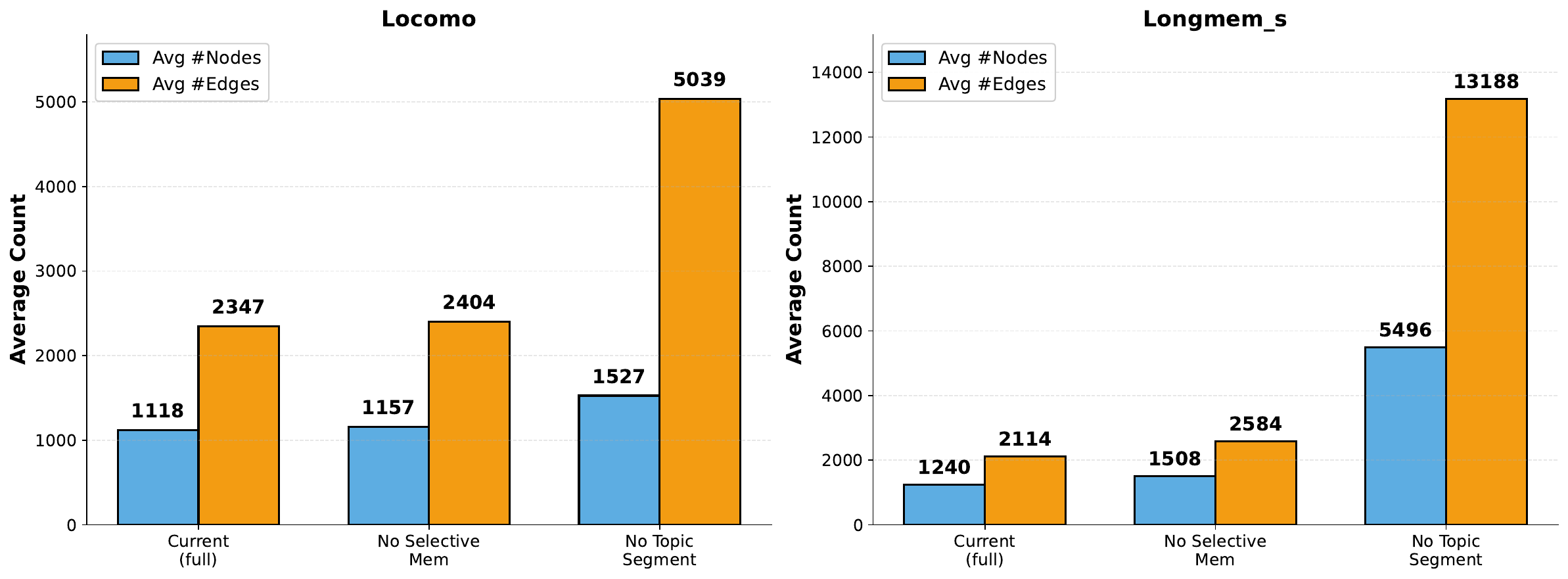}
    \caption{Graph complexity comparison across ablation configurations.}
    \label{fig:draft_graph_complexity}
\end{figure}

Figure~\ref{fig:draft_graph_complexity} shows that this configuration produces significantly denser memory graphs, whereas the full pipeline yields more compact structures. This reduction in graph complexity helps suppress irrelevant connections (reducing noise during traversal) and lowers computational overhead---consistent with the observed gains in both accuracy and efficiency.

\subsubsection{Dynamic Edge Weighting}
\label{sec:ablation_weighting}

Table~\ref{tab:ablation_weight} shows that dynamic edge weighting consistently improves retrieval across both benchmarks and granularities. For instance, turn-level R@10 increases by +1.88 on LongMemEval-s (89.75 $\rightarrow$ 91.63) and +2.67 on LOCOMO-10 (62.01 $\rightarrow$ 64.68) over uniform weighting. These gains---though modest in magnitude---are stable across settings, suggesting that query-conditioned edge weights help adapt retrieval to shifting dialogue context.

\begin{table}[H]
\centering
\caption{Dynamic Edge Weighting}
\label{tab:ablation_weight}
\small
\setlength{\tabcolsep}{4pt}
\renewcommand{\arraystretch}{1.05}
\begin{tabular}{l|cc}
\hline
\textbf{Method} & \textbf{R@10 Turn} & \textbf{R@10 Session} \\
\hline
\multicolumn{3}{c}{\textit{Locomo}} \\
\hline
Dynamic Weight & 64.68 & 92.03  \\
Uniform (w=1)  & 62.01 & 91.02  \\
\hline
\multicolumn{3}{c}{\textit{Longmem\_s}} \\
\hline
Dynamic Weight & 91.63 & 98.54  \\
Uniform (w=1)  & 89.75 & 97.70  \\
\hline
\end{tabular}
\end{table}

\subsubsection{Query-Focused Subgraph Retrieval and Triplet Enrichment}
\label{sec:ablation_retrieval_gen}

We examine two design choices: (1) restricting PageRank to a query-focused subgraph, and (2) enriching retrieved turns with their supporting knowledge graph triplets.

First, Tables~\ref{tab:fullgraph_recall} and~\ref{tab:subgraph_efficiency} compare full-graph versus subgraph-based retrieval. Subgraph retrieval consistently improves recall---e.g., +12.91 in turn-level R@10 on LOCOMO-10 (51.77 $\rightarrow$ 64.68)---while reducing PPR latency (14.19 ms $\rightarrow$ 12.44 ms) and maintaining high turn coverage (94.31\%). On LongMemEval-s, the latency gain is larger (18.34 ms $\rightarrow$ 14.21 ms), with near-perfect coverage (99.90\%). These results suggest that limiting traversal to a relevance-bounded subgraph helps suppress distant or low-signal nodes, thereby reducing noise without substantial loss of recall---particularly valuable in sparse or fragmented dialogue histories.

\begin{table}[H]
\centering
\caption{Full Graph vs Subgraph: Retrieval Performance}
\label{tab:fullgraph_recall}
\small
\setlength{\tabcolsep}{4pt}
\renewcommand{\arraystretch}{1.05}
\begin{tabular}{l|cc}
\hline
\textbf{Method} & \textbf{R@10 Turn} & \textbf{R@10 Session}  \\
\hline
\multicolumn{3}{c}{\textit{Locomo}} \\
\hline
Full Graph & 51.77 & 86.13  \\
Subgraph   & 64.68 & 92.03  \\
\hline
\multicolumn{3}{c}{\textit{Longmem\_s}} \\
\hline
Full Graph & 88.91 & 96.86  \\
Subgraph   & 91.63 & 98.54  \\
\hline
\end{tabular}
\end{table}

\begin{table}[H]
\centering
\caption{Full Graph vs Subgraph: Efficiency \& Coverage}
\label{tab:subgraph_efficiency}
\small
\setlength{\tabcolsep}{4pt}
\renewcommand{\arraystretch}{1.05}
\begin{tabular}{l|cc}
\hline
\textbf{Method} & \textbf{PPR (ms)} & \textbf{Turn Cov (\%)} \\
\hline
\multicolumn{3}{c}{\textit{Locomo}} \\
\hline
Full Graph & 14.19 & 100.0 \\
Subgraph   & 12.44 & 94.31  \\
\hline
\multicolumn{3}{c}{\textit{Longmem\_s}} \\
\hline
Full Graph & 18.34 & 100.0 \\
Subgraph   & 14.21 & 99.90  \\
\hline
\end{tabular}
\end{table}

Second, Table~\ref{tab:ablation_generation} assesses the impact of injecting retrieved triplets during generation. Augmenting turns with their associated relational context consistently improves output quality: on LOCOMO-10, GPT-4o judge scores increase from 51.66 to 60.22 (+8.45), and BLEU more than doubles (13.58 $\rightarrow$ 33.00). Larger gains are observed on LongMemEval-s (GPT-4o-J: +13.83), where answers often require precise entity or temporal grounding. This pattern indicates that structured context helps the generator resolve ambiguities---e.g., distinguishing between similarly phrased user intents or tracking evolving preferences---yielding responses that are not only more fluent but also more factually grounded.

\begin{table}[H]
\centering
\small
\setlength{\tabcolsep}{4pt}
\caption{Ablation Study: Impact of Triplet Context Enrichment on Generation}
\label{tab:ablation_generation}
\begin{tabular}{l|cccc}
\hline
\textbf{Method} & \textbf{GPT4o-J} & \textbf{F1} & \textbf{BLEU} & \textbf{R-L} \\
\hline
\multicolumn{5}{c}{\textit{Locomo}} \\
\hline
Turn only & 51.66 & 27.43 & 13.58 & 37.58 \\
Turn + Triplets & 60.22 & 51.89 & 33.00 & 45.97 \\
\hline
\multicolumn{5}{c}{\textit{Longmem\_s}} \\
\hline
Turn only & 61.72 & 49.58 & 27.43 & 49.47 \\
Turn + Triplets & 75.55 & 56.71 & 33.00 & 56.58 \\
\hline
\end{tabular}
\end{table}

\section{Conclusion}
\label{sec:conclusion}

We introduce \textbf{MemORAI}, a memory framework that integrates selective filtering with segment-level summarization to retain user-relevant content while preserving global context; a provenance-enriched heterogeneous graph linking entities, turns, and segments for fine-grained, auditable retrieval; and dynamic weighted PageRank to construct query-focused subgraphs with context-aware edge weighting for prioritizing relevant evidence. Experiments on multi-session benchmarks show consistent improvements over strong baselines in turn-level recall and factual correctness under controlled conditions. These results highlight the value of jointly modeling memory granularity, temporal provenance, and query adaptation to enhance long-horizon memory utilization, enabling more coherent and reliable extended interactions.

\section*{Limitations}
While MemORAI demonstrates strong performance on existing long-horizon benchmarks, its reliance on turn-level provenance and static entity linking may limit adaptability in highly dynamic or ambiguous conversational contexts—e.g., when user intent shifts abruptly or coreferences span many sessions with sparse explicit mentions. A primary challenge of our current framework remains the high computational overhead and memory requirements inherent in deploying large-scale LLMs and complex graph-based retrieval in real-time. These constraints limit accessibility in resource-constrained environments. To address this, knowledge distillation (KD) \citep{truong2026ctpd,hoang2026mcw,vu2026dwa} has emerged as a crucial technique to transfer capabilities from powerful teacher models to more compact architectures. In future work, we plan to optimize MemORAI for more efficient deployment by integrating Small Language Models (SLMs) and specialized small-scale embedding models. We aim to leverage advanced distillation frameworks \citep{truong2025emo,an2026mol,le2025token} to ensure that smaller models maintain high-fidelity personalized memory retrieval capabilities.

\section*{Acknowledgments}
Trung Le was supported by the Air Force Office of Scientific Research under award number FA2386-25-1-4023 and the ARC Discovery Project grant DP250100262.


\bibliography{custom}

\appendix

\section{Dataset Statistics}
\label{app:locomo-stat}

\begin{table}[t]
\centering
\setlength{\tabcolsep}{8pt}
\renewcommand{\arraystretch}{1.15}

\begin{tabularx}{\linewidth}{l>{\centering\arraybackslash}X}
\toprule
\textbf{Metric} & \textbf{LoCoMo-10} \\
\midrule
Total Conversations      & 10 \\
Avg.\ Sessions per Conv. & 27.2 \\
Avg.\ Query per Conv.    & 198.6 \\
Avg.\ Tokens per Conv.   & 20{,}078.9 \\
Session Dates Annotated  & \cmark \\
Retrieval Ground-Truth   & \cmark \\
QA Ground-Truth          & \cmark \\
Conversation Subject     & User--User \\
\bottomrule
\end{tabularx}
\caption{\label{tab:locomo-stats}
    Statistics of the \textbf{LoCoMo-10} dataset. ``Avg.'' denotes per-conversation averages.}
\end{table}

\noindent\textbf{Dataset Overview.}
\textbf{LoCoMo-10} \cite{maharana2024evaluatinglongtermconversationalmemory} is a curated subset of the larger \texttt{LoCoMo} benchmark designed for evaluating long-term conversational memory. It contains ten extended \emph{user–user dialogues}, each averaging about \textbf{27 sessions} and roughly \textbf{20k tokens}.  

Unlike assistant-style datasets, LoCoMo focuses on \emph{natural human conversation flow}, where topics evolve, reappear, and depend on long-range context. This makes it a challenging testbed for models aiming to preserve and reason over persistent memory states. Each conversation is annotated with \textbf{session timestamps}, \textbf{retrieval ground-truth}, and \textbf{QA supervision}, allowing controlled evaluation of memory construction, temporal grounding, and information recall across distant dialogue turns.  

Overall, LoCoMo-10 captures the core difficulty of \emph{multi-session coherence}—understanding entities, events, and relationships that span days or weeks of dialogue—providing a compact yet realistic benchmark for long-term memory systems like \textbf{MemORAI}.

\noindent\textbf{Setup.}
The LoCoMo-10 benchmark divides its QA task into five reasoning categories designed to test different aspects of long-term memory: \emph{Single-hop}, \emph{Multi-hop}, \emph{Temporal Reasoning}, \emph{Open-domain Knowledge}, and \emph{Adversarial} questions.  
Each type probes a distinct ability—from retrieving local facts to integrating scattered evidence across sessions or rejecting unanswerable prompts.

\noindent\textbf{Results and Analysis.}
\textbf{MemORAI} achieves the highest scores on \textbf{Single-hop reasoning}, with strong margins across all metrics (\textbf{F1} = 24.67, \textbf{ROUGE-L} = 24.1, \textbf{BERTScore} = 87.32).  
This aligns with the system's strength in grounding on precise, session-local evidence: when the query's context resides in a single dialogue window, its description-enriched retrieval ensures that the generator accesses clean and relevant spans.

Performance remains solid on \textbf{Multi-hop} and \textbf{Temporal Reasoning} questions, indicating that adaptive propagation can recover links across sessions and maintain temporal consistency.  
However, scores are slightly lower (\textbf{F1} approximates to 15–17), reflecting the intrinsic difficulty of tracking multisession dependencies in user-user dialogues where events are implicit or temporally distant.  
In \textbf{Open-domain knowledge} cases, the model's reliance on dialogue-internal context limits factual completeness, as it does not access an external knowledge source.

The \textbf{Adversarial} subset shows the lowest scores, as expected, since these questions are intentionally unanswerable and reward the model for abstention rather than generation.  
\textbf{MemORAI} still maintains reasonable precision, implying partial robustness to misleading cues.

\noindent\textbf{Discussion.}
Overall, the pattern demonstrates that \textbf{MemORAI's recall-oriented retrieval} benefits factual QA most when key information exists within reachable context windows.  
Tasks that demand aggregation or external world knowledge remain challenging, suggesting directions for future work. Nonetheless, LoCoMo-10 confirms that \textbf{broad and accurate coverage} remains the dominant factor in long-term conversational QA performance.


\section{Robustness, Efficiency, and Cost Analysis}
\label{sec:appendix_robustness}

This appendix presents supplementary experimental evidence and discussion
addressing four aspects of \textsc{MemORAI}:
(1)~robustness to backbone LLMs of varying scales and context capacities;
(2)~reliability of LLM-dependent components under structured output errors;
(3)~indexing token cost and the trade-off between accuracy and computational
expense relative to simpler retrieval approaches; and
(4)~graph scalability and long-term memory maintenance.

\begin{table*}[t]
\centering
\small
\caption{Session-level retrieval performance. LME = LongMemEval, LOCO = LOCOMO-10.}
\label{tab:session_retrieval}
\begin{tabular}{lcccccc}
\toprule
\textbf{Model}
  & \textbf{LME R@3} & \textbf{LME R@5} & \textbf{LME R@10}
  & \textbf{LOCO R@3} & \textbf{LOCO R@5} & \textbf{LOCO R@10} \\
\midrule
Contriever           & 71.06 & 81.28 & 90.00 & 49.90 & 58.26 & 71.80 \\
HippoRAG~2           & 75.53 & 84.68 & 91.28 & 56.60 & 65.06 & 78.05 \\
\textsc{MemORAI}     & \textbf{90.17} & \textbf{94.56} & \textbf{98.54}
                     & \textbf{72.05} & \textbf{81.63} & \textbf{92.03} \\
\bottomrule
\end{tabular}
\end{table*}

\begin{table*}[t]
\centering
\small
\caption{Turn-level retrieval performance.}
\label{tab:turn_retrieval}
\begin{tabular}{lcccccc}
\toprule
\textbf{Model}
  & \textbf{LME R@3} & \textbf{LME R@5} & \textbf{LME R@10}
  & \textbf{LOCO R@3} & \textbf{LOCO R@5} & \textbf{LOCO R@10} \\
\midrule
Contriever           & 45.07 & 57.02 & 69.40 & 24.36 & 30.30 & 37.75 \\
HippoRAG~2           & 27.80 & 38.85 & 53.90 & 42.43 & 52.49 & 61.81 \\
\textsc{MemORAI}     & \textbf{71.13} & \textbf{82.64} & \textbf{91.63}
                     & \textbf{42.63} & \textbf{51.97} & \textbf{64.68} \\
\bottomrule
\end{tabular}
\end{table*}

\subsection{Cross-Backbone Evaluation}
\label{subsec:cross_backbone}

A potential concern is whether the empirical gains of \textsc{MemORAI} are
specific to a single backbone model or generalize across LLMs of different
scales and native long-context capacities. To address this, we expand our
evaluation to three open-source backbones covering a substantially broader
range of scales and supported context lengths:
\texttt{Qwen3-8B} (32,768 context length),
\texttt{openai/gpt-oss-20B} (131,072 context length), and
\texttt{Qwen3-30B-A3B} (262,144 context length).
Retrieval and generation performance on the LOCOMO benchmark are reported in
Tables~\ref{tab:cross_retrieval} and~\ref{tab:cross_generation}.

\begin{table*}[t]
\centering
\small
\caption{Retrieval performance of \textsc{MemORAI} on LOCOMO across backbone
models of increasing scale and context capacity.}
\label{tab:cross_retrieval}
\begin{tabular}{lcccccc}
\toprule
\textbf{Model}
  & \textbf{Turn R@3} & \textbf{Turn R@5} & \textbf{Turn R@10}
  & \textbf{Session R@3} & \textbf{Session R@5} & \textbf{Session R@10} \\
\midrule
Qwen3-8B            & 41.00 & 52.22 & 63.51 & 64.83 & 75.18 & 87.08 \\
openai/gpt-oss-20B  & 42.63 & 51.97 & 64.68 & 72.05 & 81.63 & 92.03 \\
Qwen3-30B-A3B       & \textbf{55.34} & \textbf{64.12} & \textbf{74.13} & \textbf{77.55} & \textbf{84.81} & \textbf{92.28} \\
\bottomrule
\end{tabular}
\end{table*}

\begin{table*}[t]
\centering
\small
\caption{Generation performance of \textsc{MemORAI} on LOCOMO across backbone
models. Metrics: GPT-4o-J~F1, BLEU, ROUGE-1/2/L, and BERTScore.}
\label{tab:cross_generation}
\begin{tabular}{lcccccc}
\toprule
\textbf{Model}
  & \textbf{GPT-4o-J F1} & \textbf{BLEU}
  & \textbf{R-1} & \textbf{R-2} & \textbf{R-L} & \textbf{BERTScore} \\
\midrule
Qwen3-8B            & 50.65 & 45.58 & 29.66 & 47.15 & 35.02 & 89.52 \\
openai/gpt-oss-20B  & 60.22 & 56.71 & 33.00 & 57.90 & 42.57 & 91.71 \\
Qwen3-30B-A3B       & \textbf{63.26} & \textbf{57.82} & \textbf{34.97} & \textbf{58.88} & \textbf{44.58} & \textbf{91.80} \\
\bottomrule
\end{tabular}
\end{table*}

The results demonstrate consistent effectiveness across all tested backbones.
Notably, even the smallest backbone (\texttt{Qwen3-8B}) achieves competitive
performance and continues to outperform strong baselines, confirming that the
gains are not attributable to any single large model. Furthermore, as backbone
scale and context capacity increase, \textsc{MemORAI} yields further
improvements across both retrieval and generation metrics, indicating that the
framework scales well with stronger long-context LLMs rather than being
undermined by their native capabilities. These results support the conclusion
that observed improvements arise from the framework design itself---selective
memory filtering, provenance-enriched graph construction, and query-adaptive
retrieval---rather than from a particular backbone choice.

\subsection{Robustness to Structured Output Errors}
\label{subsec:extraction_robustness}

Several components of \textsc{MemORAI}, including selective memory filtering
and knowledge graph extraction, rely on LLM prompting to produce structured
outputs (e.g., JSON or schema-compliant extractions). A legitimate concern is
whether errors in these outputs---such as malformed JSON or schema
violations---could degrade graph quality and downstream retrieval performance.
To examine this systematically, we measure the \emph{structured output error
rate}: the proportion of LLM outputs that fail to produce valid,
schema-compliant structured responses. Results are reported across all
compared methods and both benchmarks in Table~\ref{tab:structured_error}.

\begin{table}[htbp]
\centering
\small
\caption{Structured output error rate (\%) on LOCOMO and LongMemEval.
Secom and MemGas do not rely on JSON-based structured extraction (n/a).}
\label{tab:structured_error}
\begin{tabular}{lcc}
\toprule
\textbf{Method} & \textbf{LOCOMO} & \textbf{LongMemEval} \\
\midrule
A-Mem            & 73.91 & 74.07 \\
Mem0-g           &  2.43 &  9.10 \\
Secom            &  n/a  &  n/a  \\
MemGas           &  n/a  &  n/a  \\
Mem0             & 10.39 &  7.60 \\
\midrule
\textsc{MemORAI} & \textbf{3.70} & \textbf{2.10} \\
\bottomrule
\end{tabular}
\end{table}

\textsc{MemORAI} maintains consistently low error rates across both benchmarks,
demonstrating stable structured extraction behavior in practice. By contrast,
A-Mem exhibits error rates exceeding 73\%, substantially undermining the
integrity of its knowledge graph and subsequent retrieval. The majority of
observed failures across methods arise from malformed JSON
outputs---specifically missing closing brackets or unescaped quotation
marks---that prevent extracted knowledge from being parsed correctly.

The low error rate of \textsc{MemORAI} is attributable to two design
decisions: (i)~extraction prompts are engineered with explicit schema
constraints and formatting instructions tailored to triplet and provenance
extraction; and (ii)~a schema validation step during the offline indexing
phase discards ill-formed outputs before they propagate into the knowledge
graph. Together, these measures substantially reduce formatting errors and yield reliable schema-compliant outputs, confirming that the LLM-dependent components of \textsc{MemORAI} are robust to structured extraction failures under realistic operating conditions. The efficacy of leveraging LLMs for accurately extracting multi-aspect structured information from complex text has also been validated in recent domain-specific retrieval pipelines \citep{nguyen2025improving}.

\subsection{Indexing Token Cost and Comparison with Simpler Retrieval Approaches}
\label{subsec:token_cost}

\paragraph{Token Usage Per Session.}
\textsc{MemORAI} is designed to support relational and provenance-aware memory,
which inherently requires more structured processing than lightweight
embedding-only pipelines. Constructing compressed memory units and entity-level
links involves LLM-based extraction beyond simple similarity indexing,
introducing higher token consumption during the indexing phase. We acknowledge
this trade-off and report token usage statistics per session for transparency
in Table~\ref{tab:token_usage}.

\begin{table}[htbp]
\centering
\small
\caption{Token usage per session (Indexing + Updates) for all methods on
LOCOMO and LongMemEval.}
\label{tab:token_usage}
\resizebox{\columnwidth}{!}{%
\begin{tabular}{lcccc}
\toprule
& \multicolumn{2}{c}{\textbf{LOCOMO}} & \multicolumn{2}{c}{\textbf{LongMemEval}} \\
\cmidrule(lr){2-3} \cmidrule(lr){4-5}
\textbf{Method}
  & \textbf{Input Tok.} & \textbf{Output Tok.}
  & \textbf{Input Tok.} & \textbf{Output Tok.} \\
\midrule
A-Mem            & 12065.1 & 10771.6 & 19330.0 & 5056.3  \\
Mem0-g           &   783.0 &  1298.0 &   889.6 & 1442.7  \\
Secom            &  1633.6 &   802.2 &  3331.9 &  905.7  \\
MemGas           &  1795.9 &   549.0 &  5236.1 &  965.9  \\
Mem0             &   994.6 &  1012.4 &  1174.7 & 1344.9  \\
\midrule
\textsc{MemORAI} &  9700.4 &  3627.8 & 13859.5 & 2230.1  \\
\bottomrule
\end{tabular}%
}
\end{table}

Importantly, the indexing phase in \textsc{MemORAI} is performed as an
\emph{offline or asynchronous update process}, decoupled from real-time
response generation. Consequently, indexing cost does not affect response
latency at inference time, as retrieval operates over an already-constructed
memory graph. This design is consistent with prior graph-based memory
systems~\citep{gutirrez2025ragmemorynonparametriccontinual}. Furthermore, \textsc{MemORAI} does not
rely on extremely large proprietary models: our system uses the open-source
\texttt{openai/gpt-oss-20B} backbone (3.6B active parameters), which is
substantially more accessible than systems requiring GPT-4o for memory
construction. While \textsc{MemORAI} incurs higher indexing cost than lightweight methods such as Mem0-g, it remains notably more token-efficient than other graph-heavy approaches (e.g., A-Mem), while delivering over 4\% absolute improvement in retrieval performance on LOCOMO. Managing these computational trade-offs is a pervasive challenge in LLM deployment, akin to balancing multi-cost alignments in cross-tokenizer knowledge distillation \citep{hoang2026mcw}.

\paragraph{Why Graph-Based Retrieval over Embedding-Based Approaches?}
Graph-based methods explicitly model relationships between entities and events,
enabling meaningful connections even when related facts do not directly
co-occur in the same context. This structure natively supports multi-hop
reasoning and provenance-aware retrieval across long conversation horizons.
By contrast, embedding-only approaches rely primarily on similarity signals
and do not capture relational dependencies, making complex reasoning less
reliable~\citep{gutirrez2025ragmemorynonparametriccontinual}.

This advantage is confirmed in our controlled experiment using the same
embedding backbone (Contriever), where graph-based retrieval yields large
gains at both session and turn levels (Tables~\ref{tab:session_retrieval}
and~\ref{tab:turn_retrieval}), confirming that the improvements come from
relational modeling rather than stronger embeddings alone.

\subsection{Efficiency of Dynamic Weighted PageRank}
\label{subsec:dwpr_efficiency}

The Dynamic Weighted PageRank (DWPR) mechanism modifies standard Personalized
PageRank (PPR) by incorporating query-conditioned edge weights to improve
ranking quality for semantically relevant but sparsely connected nodes. We
examine whether this modification introduces meaningful computational overhead
and whether its retrieval gains justify the added complexity over uniform-weight
PPR.

\paragraph{Latency Comparison.}
Table~\ref{tab:dwpr_combined} compares end-to-end retrieval latency (in milliseconds)
for Traditional PPR and DWPR on both benchmarks.

\begin{table}[htbp]
\centering
\small
\caption{Latency and retrieval improvement of Dynamic Weighted PageRank
vs.\ Traditional PPR.}
\label{tab:dwpr_combined}
\resizebox{\columnwidth}{!}{%
\begin{tabular}{llccc}
\toprule
\textbf{Benchmark} & \textbf{Method}
  & \textbf{Latency (ms)} & \textbf{R@10 Turn} & \textbf{R@10 Session} \\
\midrule
\multirow{2}{*}{LOCOMO}
  & Traditional PPR           & 345.46 & 62.01 & 91.02 \\
  & DW PageRank & 349.97 & 64.68 & 92.03 \\
\midrule
\multirow{2}{*}{LongMemEval}
  & Traditional PPR           & 1256.53 & 89.75 & 97.70 \\
  & DW PageRank & 1270.12 & 91.63 & 98.54 \\
\bottomrule
\end{tabular}%
}
\end{table}

The additional latency introduced by DWPR is minimal---approximately 4.5\,ms
on LOCOMO and 13.6\,ms on LongMemEval (less than 1.4\% and 1.1\% overhead,
respectively)---confirming that DWPR operates at effectively the same
computational cost as standard PPR.

\paragraph{Retrieval Improvement.}
Despite its negligible overhead, DWPR yields consistent improvements across
both benchmarks and retrieval granularities, as shown in
Table~\ref{tab:dwpr_combined}. The purpose of DWPR is not to replace PPR
with a heavier algorithm, but to introduce a lightweight modification that
improves ranking quality while preserving the efficiency of standard PPR.
These results establish a favorable cost-effectiveness profile, particularly
in long-context settings where modest retrieval improvements translate into
meaningful downstream generation gains.

\subsection{Graph Scalability and Long-Term Memory Maintenance}
\label{subsec:graph_maintenance}

A natural question for any continuously operating memory system concerns
long-term graph scalability: while Selective Filtering reduces low-utility
input at the ingestion stage, the knowledge graph itself grows under extended
deployment. We address this concern along two dimensions.

\paragraph{Structural Scalability via Incremental Updates.}
A key architectural advantage of \textsc{MemORAI} is its support for fully
incremental graph updates. Unlike tree-based retrieval approaches such as
RAPTOR~\citep{sarthi2024raptorrecursiveabstractiveprocessing}, which require recursive summarization over raw text and
must rebuild the entire tree structure whenever new information is added, our
method appends new information as nodes and edges without requiring any
re-encoding or reconstruction of the existing memory store. This design
closely mirrors LightRAG~\citep{guo2024lightrag}, which similarly adopts a modular
graph structure for efficient, localized memory updates at scale. As a result,
\textsc{MemORAI} is inherently well-suited to continual, open-ended
conversational settings where new information arrives continuously and
unpredictably.

\clearpage
\onecolumn

\section{Prompt Templates}
\label{app:prompts}
\subsection{Conversation Segmentation}
\label{app:segmentation}

\begin{center}
\begin{tcolorbox}[
    colback=gray!5!white,
    colframe=gray!40!black,
    left=2mm, right=2mm, top=2mm, bottom=2mm,
    width=\textwidth
]
\ttfamily\small
You are an expert in conversational discourse analysis. Your task is to perform topic segmentation on a multi-turn dialogue.\\

TASK DEFINITION\\
Segment the conversation into topically coherent units based on semantic relatedness. Successive conversational turns discussing the same topic should be grouped into the same segment. Create new segments when topic shifts occur.\\

INPUT FORMAT\\
A numbered sequence of conversational turns, where each message is labeled with its speaker role:\\
- Message i: [speaker]: [utterance]\\

OUTPUT FORMAT\\
Return ONLY a JSON list of lists containing integer indices: [[seg1\_indices], [seg2\_indices], ...]\\
- Each inner list represents one topical segment\\
- Indices must be integers corresponding to message numbers\\
- Segments should be non-overlapping and cover all messages\\
- Maintain chronological order\\

SEGMENTATION CRITERIA\\
- Semantic coherence: Messages within a segment should share topical focus\\
- Topic shift detection: Create new segments when conversation shifts to distinct topics\\
- Contextual continuity: Consider discourse markers and referential dependencies\\

EXAMPLE\\
INPUT CONVERSATION:\\
Message 0: user: Hey, how are you?\\
Message 1: assistant: I'm good, thanks! Just finishing up some work.\\
Message 2: user: Speaking of work, did you see the email about the project deadline?\\
Message 3: assistant: Yeah, it's been moved to next Friday.\\
Message 4: user: Okay, that gives us more time. I'll update the project plan.\\
Message 5: assistant: Perfect, thanks.\\
Message 6: user: On a different note, are you free this weekend? I was thinking of going hiking.\\
Message 7: assistant: Oh, that sounds great! I'm free on Saturday.\\

REQUIRED OUTPUT:\\
\begin{verbatim}
[[0,1,2,3,4,5],[6,7]]
\end{verbatim}

EXPLANATION: Messages 0--5 form one segment discussing work-related topics. Message 6 signals a topic shift to weekend plans, forming a new segment with message 7.\\

---\\

INPUT CONVERSATION\\
\{numbered\_messages\_str\}\\

REQUIRED OUTPUT\\
Return only the JSON list of segment indices below:\\
\end{tcolorbox}

\captionof{figure}{Conversation Segmentation}
\end{center}

\newpage
\subsection{Selective Memory Filtering}
\label{app:selective_filtering}

\begin{center}
\begin{tcolorbox}[
    colback=gray!5!white,
    colframe=gray!40!black,
    left=2mm, right=2mm, top=1mm, bottom=2mm,
    width=\textwidth
]
\ttfamily\small
You are an expert Data Curator who builds user profiles from conversations.\\

TASK\\
Analyze the conversation and return a JSON array of message indices that contain valuable information about the user.\\

WHAT COUNTS AS ``VALUABLE USER INFORMATION''\\
Keep only messages that help identify the user's personal context, including:\\
1) Personal information\\
- Biographical facts: name, age, job, education, location\\
- Possessions/ownership, experiences, achievements\\
- Relationships, life events, specific details they shared\\

2) Interests / preferences / goals\\
- Likes/dislikes, habits, goals\\
- Requests for recommendations or advice that reveal real needs\\
- Questions that reveal their situation\\

3) Contextual exchanges\\
- Questions that clarify the user's specific context\\
- Personalized suggestions the user requested\\
- Responses that explicitly reference details the user mentioned earlier\\

WHAT TO SKIP\\
- Generic knowledge not tied to this user\\
- General definitions or instructions applicable to anyone\\
- Content with no connection to the user's personal context\\

OUTPUT FORMAT\\
Return ONLY a JSON array of message indices to keep. Do not add any text outside the JSON array.\\

EXAMPLES\\

Example 1\\
{[}0{]} user: What's photosynthesis?\\
{[}1{]} assistant: Photosynthesis is the process where plants convert sunlight into energy using chlorophyll.\\
Output: [0]\\
Why keep: Message [0] indicates the user's learning interest/need. Message [1] is generic knowledge and does not add user-specific profile information.\\

Example 2\\
{[}0{]} user: My cat Luna keeps scratching the furniture\\
{[}1{]} assistant: Since Luna is scratching the furniture, try placing a scratching post near her favorite spots.\\
Output: [0, 1]\\
Why keep: Message [0] contains ownership and a specific personal detail (a cat named Luna) plus a concrete problem. Message [1] is personalized to the user's stated context.\\

Example 3\\
{[}0{]} Tom: Alex, are you moving to Berlin next week?\\
{[}1{]} Alex: Yeah. I'm moving to Berlin because I got a data engineer job. I'm worried about rent because my budget is only around 1,200 EUR/month.\\
{[}2{]} Tom: Are you going alone or with someone?\\
{[}3{]} Alex: Alone. I prefer a place near the U-Bahn so commuting is easy.\\
Output: [1, 3]\\
Why keep: [1] includes location (Berlin), job (data engineer), and a constraint/goal (rent budget). [3] adds living situation (alone) and a preference (near U-Bahn).\\

Conversation:\\
\{formatted\_conv\}\\
Output:
\end{tcolorbox}

\captionof{figure}{Selective Memory Filtering}
\end{center}

\newpage
\subsection{Segment Summarization}
\label{app:segment_summary}

\begin{center}
\begin{tcolorbox}[
    colback=gray!5!white,
    colframe=gray!40!black,
    left=2mm, right=2mm, top=2mm, bottom=2mm,
    width=\textwidth
]
\ttfamily\small
Summarize the following conversation segment into a concise summary (2--3 sentences).\\[0.5em]
Focus on:\\
- Main topic discussed\\
- Key information exchanged\\
- Important facts or decisions\\[0.5em]
Segment:\\
\{segment\_content\}\\[0.5em]
Summary:
\end{tcolorbox}

\captionof{figure}{Segment Summarization}
\end{center}

\subsection{Entity Description Extraction}
\label{app:entity_extraction}

\begin{center}
\begin{tcolorbox}[
    colback=gray!5!white,
    colframe=gray!40!black,
    left=2mm, right=2mm, top=2mm, bottom=2mm,
    width=\textwidth
]
\ttfamily\small
Based on the following conversation segment, provide a brief description for each entity in context.\\[0.5em]
IMPORTANT: For each description, cite the TURN INDICES (not message indices) where the information comes from.\\[0.5em]
Segment:\\
\{segment\}\\[0.5em]
Entities to describe: \{entity\_list\}\\[0.5em]
For each entity, write a 1--2 sentence description that captures what we learn about it in this segment, and cite the turn indices used.\\[0.5em]
Output format (one entity per line):\\
entity | description | turn\_index1,turn\_index2,...\\[0.5em]
Example output:\\
User | A software engineer leading a team of 8 at Google, working on a search ranking algorithm project with a tight deadline | 1\\
Google | The company where the user works, currently launching a new project | 1\\[0.5em]
Generate descriptions:
\end{tcolorbox}

\captionof{figure}{Entity Description Extraction}
\end{center}

\subsection{Answer Generation prompt}
\label{app:gen_prompt}

\begin{center}
\begin{tcolorbox}[
    colback=gray!5!white,
    colframe=gray!40!black,
    left=2mm, right=2mm, top=2mm, bottom=2mm,
    width=\textwidth
]
\ttfamily\small
Based on the provided conversation context and timestamps, answer the following question by adhering to these strict rules:

1. Precision: Provide the short possible answer (short phrase or single value). Use words from the context whenever possible.

2. Verification: First verify if the premise of the question matches the information in the context. If the specific detail is not mentioned or cannot be determined, strictly answer: 'The information provided is not enough'.

3. Recency: If facts conflict or change over time, rely on the most recent information provided by the user. Ignore outdated facts.

4. Temporal Reasoning: If the question involves dates or durations, calculate them accurately using the provided conversation timestamps.

5. Source Attribution: If the question asks specifically about what the Assistant or User said, quote their exact words from the conversation.

Based on the following context, answer the question.\\
\{context\}\\
Question: \{query\}\\
Answer:"""
\end{tcolorbox}

\captionof{figure}{Answer Generation prompt}
\end{center}

\subsection{Triplet Extraction with Provenance}
\label{app:triplet_extraction}

\begin{center}
\begin{tcolorbox}[
    colback=gray!5!white,
    colframe=gray!40!black,
    left=2mm, right=2mm, top=2mm, bottom=2mm,
    width=\textwidth
]
\ttfamily\small
You are a knowledge graph extractor that identifies factual statements from conversation participants.\\[0.5em]

CORE PRINCIPLES:\\
1. Extract explicitly stated information only -- avoid inference\\
2. Focus on all conversation participants equally\\
3. Capture stated facts, preferences, interests, and plans\\[0.75em]

EXTRACTION RULES:\\
1. Equal Treatment: Extract factual statements from any participant\\
2. Speaker Identification: Use the participant's identifier (username, role label, or ``Speaker[N]'')\\
3. Pronoun Resolution: Replace pronouns with the speaker's identifier\\
4. Multi-turn Tracking: If information spans multiple messages, record all relevant indices\\[0.75em]

RELATIONSHIP TYPES:\\
- Identity: is, is a, has age, is from, lives in\\
- Professional: works at, studies at, has role\\
- Preferences: likes, prefers, enjoys, is interested in\\
- Intentions: is planning to, wants to, considering\\[0.75em]

OUTPUT FORMAT:\\
entity1|relation|entity2|message\_indices\\[0.75em]

EXAMPLES:\\[0.25em]

Example 1:\\
Message 0: An: I'm a designer at Apple.\\
Message 1: Binh: I work at Microsoft as a PM.\\
Message 2: An: I love cross-company projects.\\
Output:\\
An|has role|designer|0\\
An|works at|Apple|0\\
Binh|works at|Microsoft|1\\
Binh|has role|PM|1\\
An|likes|cross-company projects|2\\[0.75em]

Example 2:\\
Message 0: Sam: I live in Tokyo.\\
Message 1: Assistant: Interesting! Do you work there too?\\
Message 2: Sam: Yes, I've been living and working in Tokyo for 2 years.\\
Output:\\
Sam|lives in|Tokyo|0,2\\
Sam|works in|Tokyo|2\\
Sam|has lived in Tokyo for|2 years|2\\[0.75em]

CONVERSATION TO ANALYZE:\\
\{segment\_text\}\\[0.5em]
Extract all factual triplets, one per line:
\end{tcolorbox}

\captionof{figure}{Triplet Extraction with Provenance}
\end{center}

\subsection{GPT-4 Judge Prompt}
\label{app:judge_prompt}

\begin{center}
\begin{tcolorbox}[
    colback=gray!5!white,
    colframe=gray!40!black,
    left=2mm, right=2mm, top=2mm, bottom=2mm,
    width=\textwidth
]
\ttfamily\small
I will give you a question, a reference answer, and a response from a model.\\
Please answer \texttt{[[yes]]} if the response contains the reference answer. Otherwise, answer \texttt{[[no]]}.\\
If the response is equivalent to the correct answer or contains all the intermediate steps to get the reference answer, you should also answer \texttt{[[yes]]}.\\
If the response only contains a subset of the information required by the answer, answer \texttt{[[no]]}.\\[0.75em]

[User Question]\\
\{question\}\\[0.5em]

[The Start of Reference Answer]\\
\{answer\}\\
{[The End of Reference Answer]}\\

[The Start of Model's Response]\\
\{response\}\\
{[The End of Model's Response]}\\

Is the model response correct? Answer \texttt{[[yes]]} or \texttt{[[no]]} only.
\end{tcolorbox}

\captionof{figure}{GPT-4 Judge Prompt}
\end{center}

\end{document}